\ifdefined\pdfoutput\pdfoutput=1\fi
\documentclass[10pt,twocolumn,letterpaper]{article}
\newif\ifArxivIncludeSupplement
\ArxivIncludeSupplementtrue
\newif\ifArxivReferenceFirstPage
\ArxivReferenceFirstPagefalse

\usepackage[pagenumbers]{cvpr}

\usepackage[T1]{fontenc}
\usepackage{placeins}
\usepackage{array}
\usepackage{colortbl}
\usepackage{multirow}
\definecolor{tableband}{gray}{0.94}

\usepackage{multicol}
\definecolor{promptheader}{HTML}{B2B2B2}
\definecolor{promptframe}{HTML}{B4B4B4}
\definecolor{promptbody}{HTML}{F8F8F8}
\newsavebox{\segmenthintbody}
\newlength{\segmenthintheight}
\newlength{\segmenthintleft}
\newlength{\segmenthintbottom}
\newlength{\segmenthinttitle}
\makeatletter
\newenvironment{segmenthintprompt}{%
  \begin{lrbox}{\segmenthintbody}%
  \begin{minipage}{\dimexpr\linewidth-10mm\relax}%
  \normalfont\fontsize{10}{13.8}\selectfont
  \setlength{\parindent}{0pt}%
  \setlength{\parskip}{6pt}%
}{%
  \end{minipage}%
  \end{lrbox}%
  \setlength{\segmenthintheight}{\dimexpr
    \ht\segmenthintbody+\dp\segmenthintbody+6mm+26pt\relax}%
  \setlength{\segmenthintleft}{5mm}%
  \setlength{\segmenthintbottom}{3mm}%
  \setlength{\segmenthinttitle}{\dimexpr\segmenthintheight-17pt\relax}%
  \setlength{\unitlength}{1pt}%
  \par\smallskip\noindent
  \begin{picture}(\strip@pt\linewidth,\strip@pt\segmenthintheight)
    \put(0,0){\includegraphics[width=\linewidth,height=\segmenthintheight]{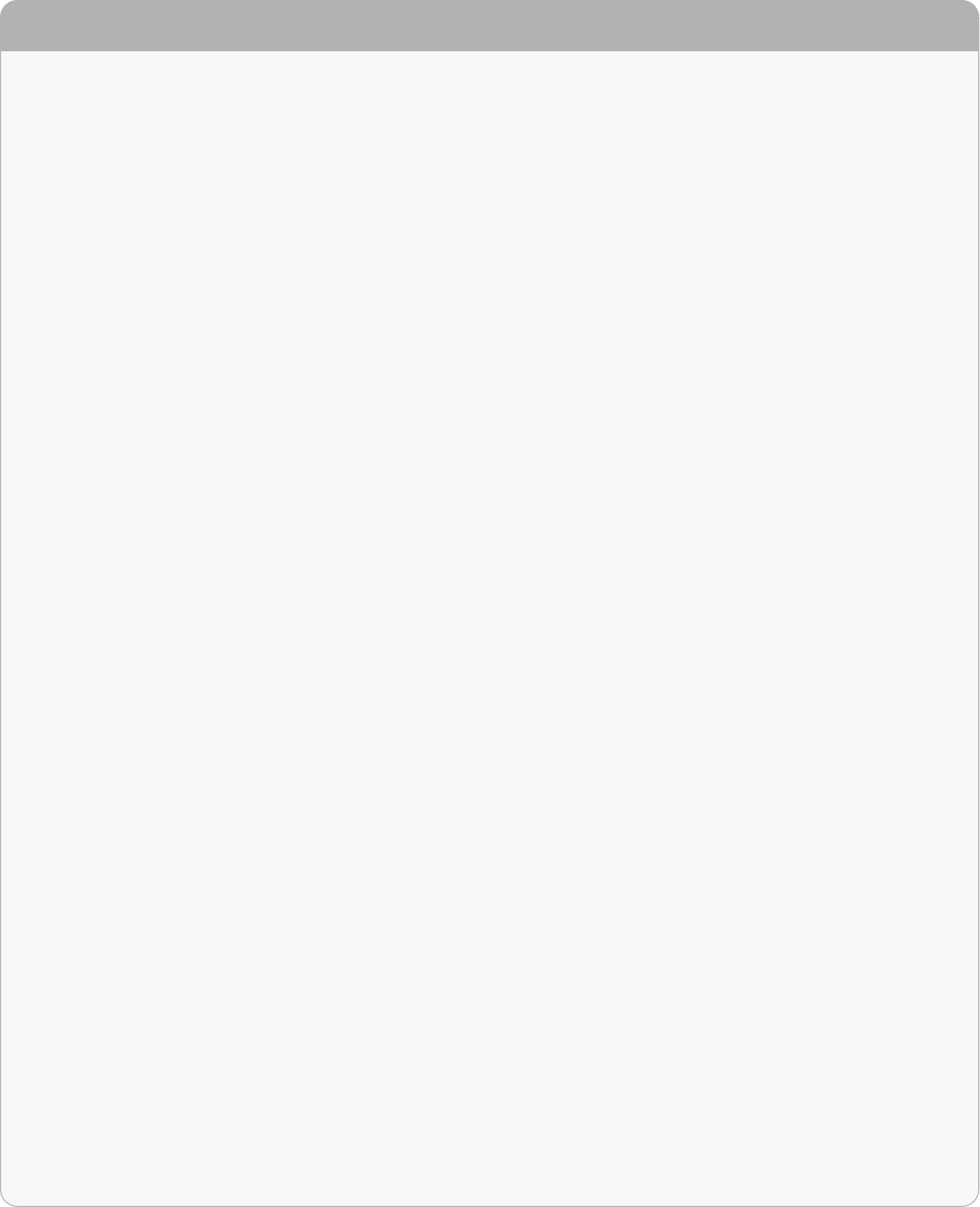}}
    \put(\strip@pt\segmenthintleft,\strip@pt\segmenthintbottom){%
      \makebox(0,0)[lb]{\usebox{\segmenthintbody}}}
    \put(\strip@pt\segmenthintleft,\strip@pt\segmenthinttitle){%
      \makebox(0,0)[lb]{\normalfont\fontsize{13}{15}\selectfont\bfseries\color{white}Prompt Template}}
  \end{picture}\par
  \ignorespacesafterend
}
\makeatother

\usepackage{microtype}

\providecommand{\nolinenumbers}{}

\definecolor{arxivlinkblue}{rgb}{0.12,0.30,0.50}
\usepackage{hyperref}
\hypersetup{
  breaklinks=true,
  colorlinks=true,
  allcolors=arxivlinkblue,
  pdftitle={Hint-Informed Self-Distillation for Reasoning Segmentation},
  pdfauthor={Kunze Jiang, Yuchen Wang, Kang Li, Qixing Zhang},
  pdfsubject={Reasoning segmentation; hint-informed self-distillation},
  pdfkeywords={reasoning segmentation, self-distillation, multimodal learning}
}

\title{Hint-Informed Self-Distillation for Reasoning Segmentation}
\author{%
  {\bfseries Kunze Jiang}\textsuperscript{1\ensuremath{\ast}},\quad
  {\bfseries Yuchen Wang}\textsuperscript{1\ensuremath{\ast}},\quad
  {\bfseries Kang Li}\textsuperscript{1},\quad
  {\bfseries Qixing Zhang}\textsuperscript{1},\\[5pt]
  \textsuperscript{1}University of Science and Technology of China\\[2pt]
  {\texttt{\{kzejiang, wangyuc\}@mail.ustc.edu.cn}}%
}
\date{}
\makeatletter
\newcommand{\ArxivAuthorFootnote}{%
  \begingroup
  \ifcsname H@@footnotetext\endcsname
    \let\@footnotetext\H@@footnotetext
  \fi
  \renewcommand{\thefootnote}{\ensuremath{\ast}}%
  \footnotetext{\fontsize{9}{11}\selectfont Equal contribution.}%
  \endgroup
}
\makeatother

\newlength{\ArxivBodyTextWidth}
\newlength{\ArxivBodyTextHeight}
\newlength{\ArxivBodyOddSideMargin}
\newlength{\ArxivBodyEvenSideMargin}
\newlength{\ArxivBodyTopMargin}
\newlength{\ArxivBodyHeadHeight}
\newlength{\ArxivBodyHeadSep}
\newlength{\ArxivBodyFootSkip}
\makeatletter
\def\ps@arxivpreprint{%
  \def\@oddhead{%
    \vbox{\hbox to\textwidth{\normalfont\normalsize Preprint\hfil}%
    \vskip3pt\hrule height.4pt}%
  }%
  \let\@evenhead\@oddhead
  \def\@oddfoot{\normalfont\normalsize\hfil\thepage\hfil}%
  \let\@evenfoot\@oddfoot
}
\renewcommand{\@maketitle}{%
  \ifArxivReferenceFirstPage\else
    \noindent{\normalfont\normalsize Preprint}\par
    \vspace{3pt}\hrule height.4pt
    \vspace{18pt}
  \fi
  \begingroup
  \centering
  {\LARGE\scshape
    Hint-Informed Self-Distillation\\
    for Reasoning Segmentation\par}
  \vspace{15pt}
  {\normalsize
      \lineskip .5em
      \begin{tabular}[t]{@{}c@{}}
        \@author
      \end{tabular}\par
  }
  \endgroup
  \vspace{8pt}
}
\newcommand{\ArxivCoverGeometry}{%
  \setlength{\textwidth}{5.5in}%
  \setlength{\textheight}{9in}%
  \setlength{\oddsidemargin}{.5in}%
  \setlength{\evensidemargin}{.5in}%
  \setlength{\topmargin}{-40pt}%
  \setlength{\headheight}{12pt}%
  \setlength{\headsep}{40pt}%
  \setlength{\footskip}{20pt}%
}
\newcommand{\ArxivRestoreBodyGeometry}{%
  \setlength{\textwidth}{\ArxivBodyTextWidth}%
  \setlength{\textheight}{\ArxivBodyTextHeight}%
  \setlength{\oddsidemargin}{\ArxivBodyOddSideMargin}%
  \setlength{\evensidemargin}{\ArxivBodyEvenSideMargin}%
  \setlength{\topmargin}{\ArxivBodyTopMargin}%
  \setlength{\headheight}{\ArxivBodyHeadHeight}%
  \setlength{\headsep}{\ArxivBodyHeadSep}%
  \setlength{\footskip}{\ArxivBodyFootSkip}%
}
\newcommand{\ArxivResetColumnHeight}{%
  \global\@colht=\textheight
  \global\@colroom=\textheight
  \global\vsize=\textheight
}
\makeatother

\ifArxivReferenceFirstPage
  \renewenvironment{abstract}{%
    \par\vspace{15pt}
    \begin{center}{\large\scshape Abstract}\end{center}
    \vspace{4pt}
    \begin{list}{}{%
      \setlength{\leftmargin}{.5in}%
      \setlength{\rightmargin}{.5in}%
      \setlength{\topsep}{0pt}%
    }%
    \item[]\normalfont\fontsize{10}{11}\selectfont\noindent\ignorespaces
  }{%
    \end{list}\par\vspace{12pt}
  }
\else
  \renewenvironment{abstract}{%
    \par
    \begin{center}{\large\scshape Abstract}\end{center}
    \vspace{5pt}
    \noindent\normalfont\ignorespaces
  }{%
    \par\vspace{12pt}
  }
\fi

\newcommand{\ArxivCoverTeaser}{%
  \begingroup
  \renewenvironment{figure}[1][]{%
    \noindent\begin{minipage}{\textwidth}%
    \centering\captionsetup{type=figure}%
  }{%
    \end{minipage}\par
  }%
  \let\ArxivCoverIncludeGraphics\includegraphics
  \renewcommand{\includegraphics}[2][]{%
    \ArxivCoverIncludeGraphics[height=3in]{##2}%
  }%
\begin{figure}[t]
  \centering
  \includegraphics[width=\linewidth]{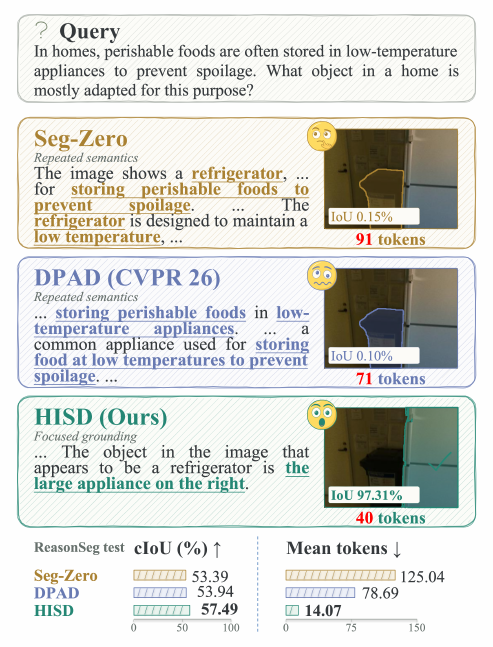}
  \caption{For the same query, HISD selects the refrigerator, while Seg-Zero
  and DPAD select the black trash bin. The charts report aggregate cIoU and
  average reasoning-token counts on the ReasonSeg test set from local runs
  under the same budget.}
  \label{fig:teaser}
\end{figure}
  \endgroup
}

\newcommand{\ArxivReadIntroduction}{%
  \ifArxivReferenceFirstPage
    \let\ArxivOriginalInput\input
    \renewcommand{\input}[1]{%
      \ifstrequal{##1}{sec/figures/1_teaser}{}{%
        \input{##1}%
      }%
    }%
    \section{Introduction}\label{sec:intro}

Reasoning segmentation requires identifying the object implicitly referred to
by a query and determining its mask amid similar objects, overlapping
attributes, and ambiguous spatial relations.
LISA~\cite{ref1} established this task, while Seg-Zero~\cite{ref2} uses
reinforcement learning to generate reasoning and geometric prompts for
segmentation.
In outcome-supervised group relative policy optimization (GRPO)~\cite{ref14},
tokens within the same response share a group-relative advantage.
This signal can reward localization outcomes but has difficulty distinguishing
tokens that support target identification from incidental descriptions,
motivating finer-grained credit assignment.

\begin{figure}[t]
  \centering
  \includegraphics[width=\linewidth]{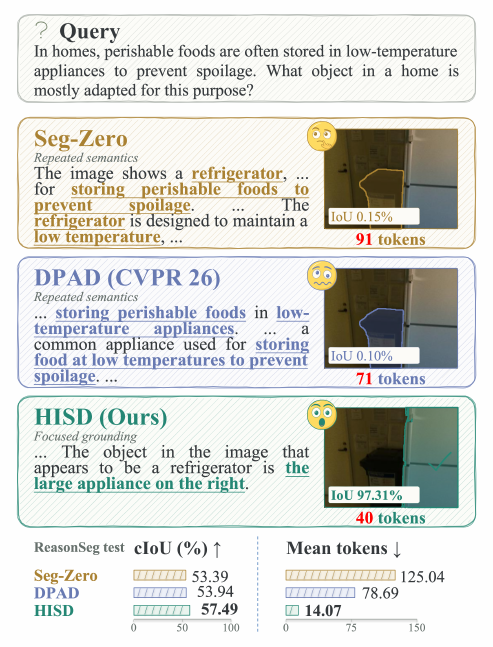}
  \caption{For the same query, HISD selects the refrigerator, while Seg-Zero
  and DPAD select the black trash bin. The charts report aggregate cIoU and
  average reasoning-token counts on the ReasonSeg test set from local runs
  under the same budget.}
  \label{fig:teaser}
\end{figure}

Two complementary lines of progress refine this training signal.
DPAD~\cite{ref3} constructs a binary reward by contrasting the relevance of an
anchored description to the target region and the full image, directly
incentivizing the distinction between the target and its context, while credit
is still assigned at the response level.
OPSD~\cite{ref29} turns verified solutions into dense supervision from teacher
distributions along the student's own trajectories, while SDPO~\cite{ref31}
extends this approach to rich environmental feedback.
Self-distillation thus makes the information available to the teacher a key
design choice.
For reasoning segmentation, this information must link the query to visible
differences among candidate objects and translate these differences into
token-level guidance.

SWiM~\cite{ref33} provides such guidance through online working memory: student
responses ranked by segmentation quality retain complete reasoning and
localization information, and a memory-conditioned teacher provides
Jensen-Shannon divergence supervision over all sampled responses, jointly
optimized with GRPO.
It can reuse continually updated experience and also supports memory-free inference,
with its guidance evolving with the student's sampled attempts.
We explore a complementary source: query-relevant visual evidence generated
from annotated targets.
This choice links the content of the teacher's evidence to its role in
optimization, using observations that support target disambiguation to
modulate how strongly tokens are reinforced.

We propose Hint-Informed Self-Distillation (HISD) to establish this link.
An external model generates Segment Hints offline from existing training
images, queries, and annotated target bounding boxes, without requiring each
hint to be written manually.
Each hint organizes target identity, visible cues, confusable objects, and a
path from an anchor to the target into query-conditioned localization evidence.
Following RLSD~\cite{ref32}, we preserve the reward-determined sign of the
advantage and modulate its magnitude using a clipped, stop-gradient
teacher-to-student evidence ratio.
RLSD applies direction-aware reweighting to both positive and negative
advantages; HISD uses hints for reinforcement under positive advantages while
retaining the base update for nonpositive advantages, reflecting the intended
use of hints to support target-oriented reinforcement.
During training, the teacher rescores student trajectories with hint context;
the student takes the image and query as input and makes predictions
independently at inference.

With the same backbone model and training data, HISD improves segmentation
accuracy on ReasonSeg and the RefCOCO family.
Compared with our reproduction of DPAD, ReasonSeg test gIoU and cIoU improve
by 3.00 and 3.55 percentage points, respectively; the average token count of
valid reasoning spans decreases by 80.17--91.83\% across the five
evaluated splits.
Table~\ref{tab:hisd_ablation} shows that, in the tested settings, Segment Hint
outperforms answer-only teacher inputs and teacher inputs with Gaussian blur
outside the bounding box, and answer-only guidance can also improve performance
on ReasonSeg.
The reported input and reweighting ablations support combining task-relevant
evidence with constrained credit weighting.
Figure~\ref{fig:teaser} shows the corresponding accuracy and reasoning-length
results.
Our contributions are as follows:

\begin{itemize}[leftmargin=*,nosep]
    \item \textbf{Self-distillation with privileged segmentation hints.} We propose an online self-distillation method in which the teacher receives additional segmentation hints, while the student receives only the image and the question. On successful and parseable trajectories, the teacher rescales only the scalar magnitude of the advantage.
    \item \textbf{Disambiguation-oriented evidence.} We automatically generate query-relevant Segment Hints from existing target annotations, without manually written hint text.
    \item \textbf{Evidence on accuracy and length.} Under a unified protocol, we demonstrate improved segmentation accuracy and shorter reasoning segments, and analyze teacher inputs and credit weighting through ablations.
\end{itemize}
    \let\input\ArxivOriginalInput
  \else
    \section{Introduction}\label{sec:intro}

Reasoning segmentation requires identifying the object implicitly referred to
by a query and determining its mask amid similar objects, overlapping
attributes, and ambiguous spatial relations.
LISA~\cite{ref1} established this task, while Seg-Zero~\cite{ref2} uses
reinforcement learning to generate reasoning and geometric prompts for
segmentation.
In outcome-supervised group relative policy optimization (GRPO)~\cite{ref14},
tokens within the same response share a group-relative advantage.
This signal can reward localization outcomes but has difficulty distinguishing
tokens that support target identification from incidental descriptions,
motivating finer-grained credit assignment.

\begin{figure}[t]
  \centering
  \includegraphics[width=\linewidth]{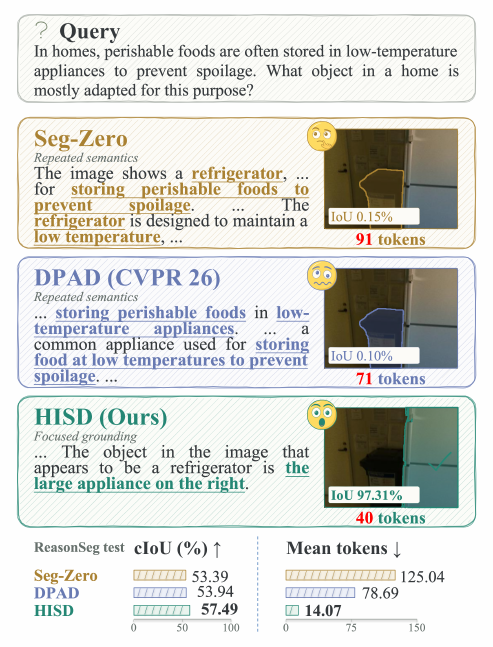}
  \caption{For the same query, HISD selects the refrigerator, while Seg-Zero
  and DPAD select the black trash bin. The charts report aggregate cIoU and
  average reasoning-token counts on the ReasonSeg test set from local runs
  under the same budget.}
  \label{fig:teaser}
\end{figure}

Two complementary lines of progress refine this training signal.
DPAD~\cite{ref3} constructs a binary reward by contrasting the relevance of an
anchored description to the target region and the full image, directly
incentivizing the distinction between the target and its context, while credit
is still assigned at the response level.
OPSD~\cite{ref29} turns verified solutions into dense supervision from teacher
distributions along the student's own trajectories, while SDPO~\cite{ref31}
extends this approach to rich environmental feedback.
Self-distillation thus makes the information available to the teacher a key
design choice.
For reasoning segmentation, this information must link the query to visible
differences among candidate objects and translate these differences into
token-level guidance.

SWiM~\cite{ref33} provides such guidance through online working memory: student
responses ranked by segmentation quality retain complete reasoning and
localization information, and a memory-conditioned teacher provides
Jensen-Shannon divergence supervision over all sampled responses, jointly
optimized with GRPO.
It can reuse continually updated experience and also supports memory-free inference,
with its guidance evolving with the student's sampled attempts.
We explore a complementary source: query-relevant visual evidence generated
from annotated targets.
This choice links the content of the teacher's evidence to its role in
optimization, using observations that support target disambiguation to
modulate how strongly tokens are reinforced.

We propose Hint-Informed Self-Distillation (HISD) to establish this link.
An external model generates Segment Hints offline from existing training
images, queries, and annotated target bounding boxes, without requiring each
hint to be written manually.
Each hint organizes target identity, visible cues, confusable objects, and a
path from an anchor to the target into query-conditioned localization evidence.
Following RLSD~\cite{ref32}, we preserve the reward-determined sign of the
advantage and modulate its magnitude using a clipped, stop-gradient
teacher-to-student evidence ratio.
RLSD applies direction-aware reweighting to both positive and negative
advantages; HISD uses hints for reinforcement under positive advantages while
retaining the base update for nonpositive advantages, reflecting the intended
use of hints to support target-oriented reinforcement.
During training, the teacher rescores student trajectories with hint context;
the student takes the image and query as input and makes predictions
independently at inference.

With the same backbone model and training data, HISD improves segmentation
accuracy on ReasonSeg and the RefCOCO family.
Compared with our reproduction of DPAD, ReasonSeg test gIoU and cIoU improve
by 3.00 and 3.55 percentage points, respectively; the average token count of
valid reasoning spans decreases by 80.17--91.83\% across the five
evaluated splits.
Table~\ref{tab:hisd_ablation} shows that, in the tested settings, Segment Hint
outperforms answer-only teacher inputs and teacher inputs with Gaussian blur
outside the bounding box, and answer-only guidance can also improve performance
on ReasonSeg.
The reported input and reweighting ablations support combining task-relevant
evidence with constrained credit weighting.
Figure~\ref{fig:teaser} shows the corresponding accuracy and reasoning-length
results.
Our contributions are as follows:

\begin{itemize}[leftmargin=*,nosep]
    \item \textbf{Self-distillation with privileged segmentation hints.} We propose an online self-distillation method in which the teacher receives additional segmentation hints, while the student receives only the image and the question. On successful and parseable trajectories, the teacher rescales only the scalar magnitude of the advantage.
    \item \textbf{Disambiguation-oriented evidence.} We automatically generate query-relevant Segment Hints from existing target annotations, without manually written hint text.
    \item \textbf{Evidence on accuracy and length.} Under a unified protocol, we demonstrate improved segmentation accuracy and shorter reasoning segments, and analyze teacher inputs and credit weighting through ablations.
\end{itemize}
  \fi
}

\newcommand{\ArxivBeginBody}{%
  \ifArxivReferenceFirstPage
    \clearpage
    \ArxivRestoreBodyGeometry
    \pagestyle{plain}
    \twocolumn
    \ArxivResetColumnHeight
  \fi
  \flushbottom
}
\AtBeginDocument{%
  \ifArxivReferenceFirstPage
    \ArxivCoverGeometry
    \onecolumn
    \pagestyle{arxivpreprint}
    \raggedbottom
  \else
    \pagestyle{plain}
  \fi
}

\begin{document}
\maketitle
\ifArxivReferenceFirstPage
  \thispagestyle{arxivpreprint}
\fi
\ArxivAuthorFootnote
\begin{abstract}
Reasoning segmentation requires identifying the target referred to by a query
among visually similar objects, yet trajectory-level rewards cannot directly
indicate which tokens provide relevant visual evidence.
We propose Hint-Informed Self-Distillation (HISD), which uses
target-disambiguation evidence for token-level credit assignment.
An external model generates Segment Hints offline from training images,
queries, and ground-truth target annotations, organizing target identity,
discriminative attributes, confusable objects, and anchoring paths.
A hint-conditioned exponential moving average teacher scores the student's
own trajectories, and HISD computes the teacher-to-student evidence ratio
over a token support set to adjust only the magnitudes of positive
group-relative advantages, leaving nonpositive advantages unchanged.
This design encourages concise reasoning while retaining the information
needed for target disambiguation, and the student requires neither hints
nor a teacher at inference.
Under a unified training protocol, HISD improves ReasonSeg test gIoU and
cIoU over our reproduction of DPAD by 3.00 and 3.55 percentage
points, respectively.
Across the five evaluated splits of ReasonSeg and the RefCOCO family, HISD
achieves higher segmentation metrics than this baseline and reduces the
average token count of valid reasoning spans by 80.17--91.83\%.
\end{abstract}

\ifArxivReferenceFirstPage
  \ArxivCoverTeaser
\fi
\ArxivBeginBody
\ArxivReadIntroduction
\section{Related Work}\label{sec:related_work}

\subsection{Reasoning Segmentation}

Reasoning segmentation links query understanding to mask prediction, a task
paradigm established by LISA~\cite{ref1}.
Methods such as Seg-Zero~\cite{ref2}, SAM-R1~\cite{ref5}, Seg-R1~\cite{ref6},
and VisionReasoner~\cite{ref4} use outcome rewards in reinforcement learning
to improve visual localization, reducing reliance on annotated reasoning
processes.
However, geometric outcomes still provide only indirect supervision for
disambiguation: trajectory-level scores cannot distinguish tokens that rule
out distractors from incidental narration.
DPAD~\cite{ref3} introduces semantic discrimination by comparing the
CLIP~\cite{ref44} similarity of generated anchored descriptions to ground-truth
target-region crops and to the full image.
Its binary reward encourages target-oriented reasoning and complements
geometric rewards, but still acts on the entire trajectory rather than
providing token-specific credit.
Thus, how visual disambiguation evidence should influence individual token
updates remains to be specified.

\subsection{\texorpdfstring{\raggedright Reinforcement Learning with \mbox{Privileged Teachers}}{Reinforcement Learning with Privileged Teachers}}

Methods that use privileged context differ both in their sources of evidence
and in how the evidence enters optimization.
OPSD~\cite{ref29} uses verified reasoning trajectories, while
SDPO~\cite{ref31} uses rich environmental feedback; both distill conditioned
next-token distributions along student-sampled trajectories, providing denser
supervision than scalar outcomes.
SWiM~\cite{ref33} constructs such context online from complete student
trajectories selected by segmentation quality.
This self-generated memory contains reasoning processes and localization
proposals; teacher and student distributions are aligned through JSD over all
sampled trajectories and jointly optimized with GRPO~\cite{ref14}.
RLSD~\cite{ref32} instead scales reward-derived advantages using a
stop-gradient teacher-to-student evidence ratio, with clipping and mixing.
Using this direction--magnitude separation framework, HISD studies
how query-relevant disambiguation content informs
token-level credit assignment.

\section{Method}\label{sec:method}

\begin{figure*}[t]
  \centering
  \includegraphics[width=\linewidth]{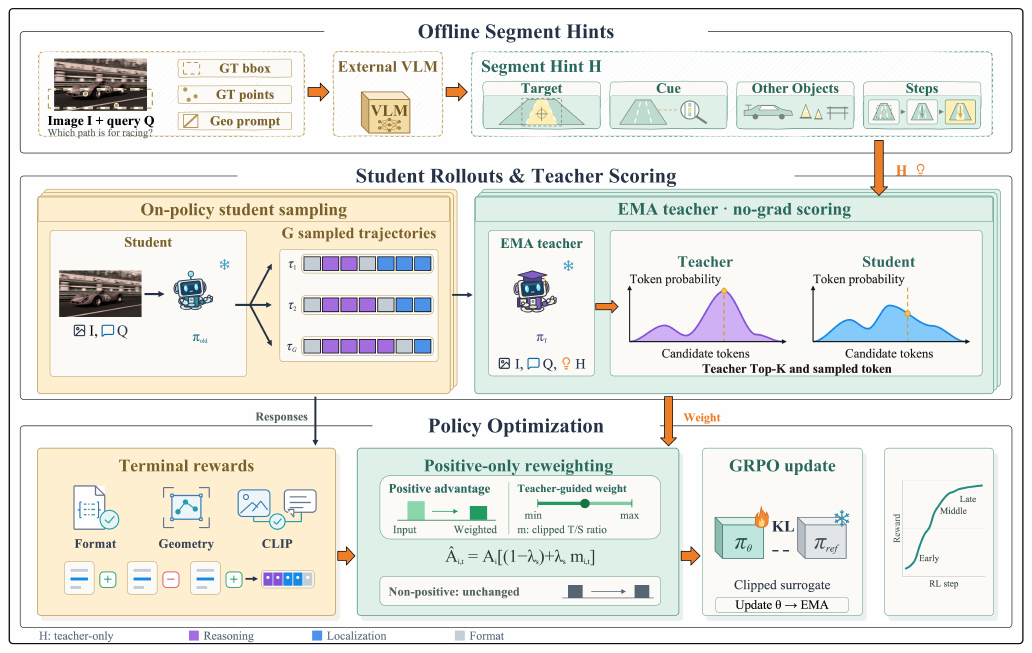}
  \caption{Hint-Informed Self-Distillation (HISD) pipeline.
  Offline Segment Hints condition an exponential moving average (EMA)
  teacher that scores the same on-policy student responses.
  Terminal rewards supply trajectory advantages; teacher-derived weights
  modulate positive-advantage updates, while nonpositive advantages retain
  the original reward signal. The token sequences, probability curves,
  and reward curve are schematic. The hint and teacher are used only during training.}
  \label{fig:pipeline}
\end{figure*}

\subsection{HISD Framework Overview}\label{sec:hisd_overview}

We propose Hint-Informed Self-Distillation (HISD), a structured self-distillation framework for reasoning
segmentation.
Figure~\ref{fig:pipeline} summarizes this setup: during training,
an exponential moving average (EMA) teacher conditioned on the Segment Hint $H$
rescores the same completed
student trajectory, while at test time the student receives only the image
and query.
Formally, given an image--query pair $x=(I,Q)$, the trainable
vision--language policy $\pi_\theta$ autoregressively generates a response
$y=(y_1,\ldots,y_{|y|})$, i.e., $y\sim\pi_\theta(\cdot\mid x)$.
For a group of $G$ sampled responses, the trajectory-level advantage is
\begin{equation}
  A^{(i)} = \frac{R(x,y^{(i)})-\bar{R}}{\sigma_R},
  \qquad
  \bar{R} = \frac{1}{G}\sum_{j=1}^{G}R(x,y^{(j)}).
  \label{eq:trajectory_advantage}
\end{equation}
Here, $R$ is the environment reward, and $\sigma_R$ is its within-group
standard deviation.
This reward is inherently sparse: it indicates only
whether the entire response succeeds and cannot identify which tokens perform
target disambiguation and spatial anchoring.
To address this limitation, HISD
adds two auxiliary components.
First, a teacher conditioned on privileged
information provides token-level evaluations along trajectories already
generated by the student.
Second, an external model generates the privileged
information offline from the query, image, and ground-truth box.
The teacher distribution is defined as
\begin{equation}
  P_{\mathrm{T}}(y_t\mid x,H,y_{<t})
  = \pi_{\bar{\theta}}(y_t\mid I,Q\oplus H,y_{<t}).
  \label{eq:teacher_distribution}
\end{equation}
Here, $H$ denotes the Segment Hint, $\oplus$ denotes text concatenation, and
$\bar{\theta}$ denotes the EMA teacher parameters
(Section~\ref{sec:stable_optimization}).

For advantage reweighting, we apply the weight only to positive-advantage
trajectories; negative-advantage trajectories retain their original GRPO
advantages.
The piecewise definition is given in
Section~\ref{sec:positive_reweighting}.

The overall training objective retains the policy-gradient form
\begin{equation}
  \begin{aligned}
  \nabla_{\theta}\mathcal{J}(\theta)
  ={}& \mathbb{E}_{x,y\sim\pi_{\theta}(\cdot\mid x)}\!\Bigl[
    \\
    &\frac{1}{|y|}\sum_{t=1}^{|y|}
    \hat{A}_t\nabla_{\theta}\log\pi_{\theta}(y_t\mid x,y_{<t})
  \Bigr].
  \end{aligned}
  \label{eq:policy_gradient}
\end{equation}
Here, $\hat{A}_t$ is the reweighted token-level advantage defined in
Section~\ref{sec:positive_reweighting}.

\subsection{Structured Segment Hint Generation}\label{sec:hint_generation}

The design of the Segment Hint is based on one observation: failures in
reasoning segmentation may arise from misunderstanding the target, misusing
cues, mistaking a distractor for the target, or inconsistency between the
reasoning chain and the final localization.
Even in a successful response,
only some tokens actually perform target disambiguation and spatial anchoring.
Distinguishing these cases is a prerequisite for fine-grained credit
assignment.

Given a training sample $x=(I,Q)$ and generic target annotation
$B^{\mathrm{gt}}$, the external hint generator $\mathcal{G}$ produces the
structured Segment Hint offline under $P_{\mathrm{gen}}$ from the query,
image, and annotation, instantiated here as a ground-truth box supplemented
with two positive interior points:
\begin{equation}
  H
  = \mathcal{G}\!\left(I,Q,B^{\mathrm{gt}};P_{\mathrm{gen}}\right)
  = \left(h^{\mathrm{tgt}},h^{\mathrm{cue}},
    h^{\mathrm{oth}},h^{\mathrm{step}}\right).
  \label{eq:hint_generation}
\end{equation}
The field $h^{\mathrm{tgt}}$ specifies the target noun,
$h^{\mathrm{cue}}$ gives its key attributes, $h^{\mathrm{oth}}$ lists
distractors in the scene and their distinguishing attributes, and
$h^{\mathrm{step}}$ describes a reasoning path from anchors to the target.
The hint $H$ is generated offline once per sample and then serves as fixed
teacher context for that sample.

\subsection{Direction--Magnitude-Decoupled Advantage Reweighting}
\label{sec:positive_reweighting}

Given a completed trajectory, we compute the discrepancy between the teacher
and student distributions at each position over a Top-$K$ support. The support
contains the teacher's Top-$K$ tokens and the sampled token $y_t$;
$\widetilde{P}_{\mathrm{T}}$ and $\widetilde{P}_{\mathrm{S}}$ denote the
distributions renormalized to sum to one over this support. The student
signal distribution $P_{\mathrm{S}}(\cdot\mid x,y_{<t})=
\pi_\theta(\cdot\mid x,y_{<t})$ is obtained without gradients. We compute
\begin{equation}
  \begin{aligned}
  \Delta_t = \operatorname{sg}\!\big(&
    \log\widetilde{P}_{\mathrm{T}}(y_t\mid x,H,y_{<t}) \\
    &-\log\widetilde{P}_{\mathrm{S}}(y_t\mid x,y_{<t})\big).
  \end{aligned}
  \label{eq:teacher_student_gap}
\end{equation}
Here, $\operatorname{sg}$ denotes the stop-gradient operator.

HISD incorporates this signal into policy optimization by decoupling direction
and magnitude. We define a token-level reweighting factor with symmetric
clipping:
\begin{equation}
  m_t^{(i)}
  = \operatorname{clip}\!\left(
    \exp(\Delta_t^{(i)}),\,1-\epsilon_w,\,1+\epsilon_w
  \right).
  \label{eq:token_weight}
\end{equation}
Here, $\epsilon_w$ controls the weight clipping range. We apply this weight
only to positive-advantage trajectories, while negative-advantage trajectories
retain their original GRPO advantages. The token-level advantage is defined
piecewise as
\begin{equation}
  \hat{A}_t^{(i)} =
  \begin{cases}
    A^{(i)}\big[(1-\lambda)+\lambda m_t^{(i)}\big],
      & A^{(i)}>0,\\[2pt]
    A^{(i)}, & A^{(i)}\leq 0.
  \end{cases}
  \label{eq:corrected_advantage}
\end{equation}
We reweight only positive advantages because $H$ describes a path to the
correct target rather than the cause of error in a particular failed
trajectory. On a negative-advantage trajectory, the teacher can judge only
whether a token is consistent with the hinted path. Strengthening suppression
on this basis would mistake inconsistency with the hint for an error.

\subsection{Stable Policy Optimization}\label{sec:stable_optimization}

Token-level credit assignment over long reasoning chains can amplify training
instability. HISD stabilizes optimization through three design choices: curriculum
scheduling, an EMA teacher, and a clipped surrogate objective.

\paragraph{Curriculum decay.}
Early in training, teacher-guided token-level reweighting modulates the
updates; as training progresses, the influence of self-distillation is
gradually annealed. The mixing coefficient $\lambda$ in
Eq.~\eqref{eq:corrected_advantage} decays linearly with optimizer steps.
Let $s$ denote the current optimizer step and $T$ the total number of
optimization steps. The schedule is
\begin{equation}
  \lambda_s
  = \lambda_0\cdot\max\!\left(0,\,1-\frac{s}{\kappa T}\right).
  \label{eq:curriculum_schedule}
\end{equation}
Here, $\lambda_0$ is the initial coefficient and $\kappa T$ is the decay
duration.

\paragraph{EMA teacher.}
To stabilize the teacher signal, we maintain an exponential moving average of
the policy parameters as the teacher parameters. Following Mean
Teacher~\cite{ref28}, we update them after every optimizer step:
\begin{equation}
  \bar{\theta}
  \leftarrow (1-\beta)\bar{\theta}+\beta\theta.
  \label{eq:ema_update}
\end{equation}
Here, $\beta$ is the student mixing rate.

\paragraph{Token-wise GRPO clipped objective.}
Finally, we optimize the policy with a GRPO-style clipped surrogate objective.
For token $t$ in trajectory $i$, let $\rho_t^{(i)}$ be the importance ratio
between the current policy and the old policy used for sampling. The
token-wise clipped term is
\begin{equation}
  \begin{aligned}
  \ell_t^{(i)}(\theta)
  ={}& \min\!\Big(
    \rho_t^{(i)}\hat{A}_t^{(i)},
    \\
    &\quad
    \operatorname{clip}\!\left(
      \rho_t^{(i)},1-\epsilon_{\mathrm{pg}},1+\epsilon_{\mathrm{pg}}
    \right)\hat{A}_t^{(i)}
  \Big) \\
  &-\eta_{\mathrm{KL}}k_t^{(i)}
  +\eta_{\mathrm{H}}\mathcal{H}_t^{(i)}(\theta).
  \end{aligned}
  \label{eq:grpo_token_objective}
\end{equation}
Here, $\epsilon_{\mathrm{pg}}$ is the policy-ratio clipping range.
The supplementary material provides the complete training procedure in
Algorithm~1 and summarizes the hyperparameters in Table~A1.

\FloatBarrier
\section{Experiments}
\label{sec:experiments}
\renewcommand{\dblfloatpagefraction}{0.90}
\makeatletter
\setlength{\@dblfptop}{0pt}
\setlength{\@dblfpsep}{12pt plus 1fil}
\setlength{\@dblfpbot}{0pt}
\makeatother

\subsection{Experimental Setup}
\label{sec:experimental_setup}

\paragraph{Data and comparison protocol.}
Using the same set of 3,000 training examples from
RefCOCOg~\cite{ref45}, we train Seg-Zero~\cite{ref2}, DPAD~\cite{ref3},
and HISD for one epoch. All three methods use
Qwen2.5-VL-7B-Instruct~\cite{ref43} as the trainable policy and a frozen
SAM2-Large~\cite{ref42} as the deterministic segmentation module. We
evaluate on a range of standard benchmarks for reasoning segmentation
and referring expression segmentation.

\subsection{Evaluation Metrics}
\label{sec:evaluation_metrics}
The gIoU metric first computes the intersection over union between the
prediction and annotation for each image, then averages these values
across all images. The cIoU metric separately sums the intersection areas
and union areas across all images, then divides the total intersection
area by the total union area.

In the diagnostic text--region alignment analysis, $S_1$ and $S_2$ denote
the mean cosine similarities between anchored descriptions and the
ground-truth bounding-box crop and the full image, respectively. SNR is
the mean of the per-example ROI/AOI similarity ratios. $TS_1$, $TS_2$,
and TSNR replace the anchored descriptions with reasoning text. Thus,
SNR and TSNR average per-example ratios rather than dividing the mean
similarities.

\subsection{Implementation Details}
\label{sec:implementation_details}
HISD uses Gemini 2.5 Pro~\cite{ref46} to generate Segment Hints. The
teacher is initialized from the student and updated after every
optimizer step by mixing in $\beta=0.05$ of the current student
parameters (Eq.~\eqref{eq:ema_update}). We use the teacher's Top-$K$
support set with $K=16$ and symmetric weight clipping with
$\epsilon_w=0.1$. The mixing coefficient starts at $\lambda_0=0.5$ and
decays linearly to zero over the first 25\% of training steps.

\subsection{Experimental Results}
\label{sec:main_results}

\begin{table}[t]
  \centering
  \caption{ReasonSeg results (\%). $\dagger$: Results reproduced by us.}
  \label{tab:reasonseg_main}
  \fontsize{8}{9.2}\selectfont
  \renewcommand{\arraystretch}{1.08}
  \setlength{\tabcolsep}{2.2pt}
  \begin{tabular*}{\columnwidth}{@{\extracolsep{\fill}}lrrrr@{}}
    \toprule
    Method & \multicolumn{2}{c}{Val} & \multicolumn{2}{c}{Test} \\
    \cmidrule(lr){2-3}\cmidrule(l){4-5}
    & gIoU & cIoU & gIoU & cIoU \\
    \midrule
    OVSeg~\cite{ovseg} & 28.50 & 18.60 & 26.10 & 20.80 \\
    ReLA / GRES~\cite{gres} & 22.40 & 19.90 & 21.30 & 22.00 \\
    Grounded-SAM~\cite{grounded_sam} & 26.00 & 14.50 & 21.30 & 16.40 \\
    LISA-7B~\cite{ref1} & 53.60 & 52.30 & 48.70 & 48.80 \\
    LISA-13B~\cite{ref1} & 57.70 & 60.30 & 53.80 & 50.80 \\
    SAM4MLLM~\cite{sam4mllm} & 46.70 & 48.10 & -- & -- \\
    Qwen2.5-VL-3B + SAM2~\cite{pixelthink} & 53.80 & 44.10 & 47.60 & 37.40 \\
    SegZero$^{\dagger}$~\cite{ref2} & 60.05 & 53.97 & 56.36 & 53.39 \\
    DPAD$^{\dagger}$~\cite{ref3} & 60.45 & 57.38 & 57.45 & 53.94 \\
    \textbf{HISD (Ours)} & \textbf{63.76} & \textbf{61.98} & \textbf{60.45} & \textbf{57.49} \\
    \bottomrule
  \end{tabular*}
\end{table}

\begin{figure*}[p]
  \centering
  \includegraphics[width=\linewidth]{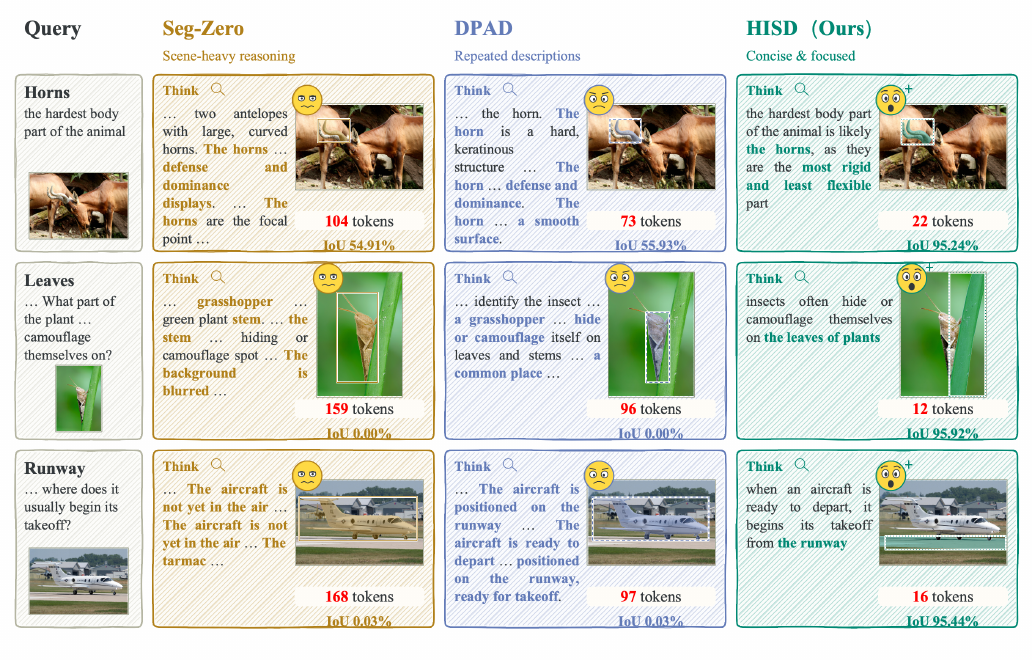}
  \caption{Comparison on horns, plant leaves, and runway queries. Predictions and verbatim output excerpts highlight HISD's target cues. Values are per-sample IoU and full reasoning-token counts. The leaves and runway queries are abridged; ellipses mark omissions.}
  \label{fig:qualitative}
\end{figure*}

\paragraph{Reasoning segmentation.}
HISD achieves the highest scores on all four ReasonSeg metrics reported
in Table~\ref{tab:reasonseg_main}. Relative to DPAD reproduced by us, it
improves validation gIoU/cIoU by 3.31/4.60 percentage points and test
gIoU/cIoU by 3.00/3.55 percentage points. The examples in
Figure~\ref{fig:qualitative} illustrate two types of improvement
consistent with these gains: avoiding the selection of distractor
objects as targets, and obtaining more accurate localization when the
target is correctly named.

\begin{table}[t]
  \centering
  \caption{cIoU (\%) on the RefCOCO-family datasets. All three evaluation splits are shown. $\dagger$: Results reproduced by us.}
  \label{tab:refcoco_main}
  \fontsize{8}{9.2}\selectfont
  \renewcommand{\arraystretch}{1.08}
  \setlength{\tabcolsep}{2.2pt}
  \begin{tabular*}{\columnwidth}{@{\extracolsep{\fill}}lccc@{}}
    \toprule
    Method & \shortstack{RefCOCO\\testA}
      & \shortstack{RefCOCO+\\testA} & \shortstack{RefCOCOg\\test} \\
    \midrule
    LAVT~\cite{lavt} & 75.80 & 68.40 & 62.10 \\
    ReLA / GRES~\cite{gres} & 76.50 & 71.00 & 66.00 \\
    LISA-7B~\cite{ref1} & 76.50 & 67.40 & 68.50 \\
    PixelLM-7B~\cite{pixellm} & 76.50 & 71.70 & 70.50 \\
    MagNet~\cite{magnet} & 78.30 & 73.60 & 69.30 \\
    SegZero$^{\dagger}$~\cite{ref2} & 74.07 & 71.59 & 68.72 \\
    DPAD$^{\dagger}$~\cite{ref3} & 79.12 & 75.02 & 71.24 \\
    \textbf{HISD (Ours)} & \textbf{80.10} & \textbf{76.31} & \textbf{73.24} \\
    \bottomrule
  \end{tabular*}
\end{table}

\paragraph{Referring expression segmentation.}
HISD improves both metrics across all three referring expression
splits, with cIoU shown in Table~\ref{tab:refcoco_main}. Relative to the
DPAD implementation, its gIoU/cIoU gains are 1.11/0.98 percentage points
on RefCOCO testA, 1.41/1.29 percentage points on RefCOCO+ testA, and
1.57/2.00 percentage points on RefCOCOg test. The gains on the RefCOCO
family are smaller than those on ReasonSeg, possibly because referring
expressions are short and direct, making target disambiguation less
demanding. The consistent improvements across both types of datasets
indicate that the trained policy can retain gains on these evaluated
expression distributions.


\paragraph{Stratified performance and reasoning length.}
We assign query types using a frozen Qwen2.5-VL-72B
image-and-expression classifier, with manual adjudication of 62 Logic
examples, and use all 779 official ReasonSeg-Diff difficulty labels
released by PixelThink~\cite{pixelthink}. All three methods are
evaluated using the same fixed labels.
Table~\ref{tab:reasonseg_stratified} shows that HISD has higher gIoU
than DPAD in every stratum while using only 11.04--15.52 reasoning
tokens on average. Its mean reasoning lengths in the easy, medium, and
hard groups are 13.24, 13.76, and 15.52 tokens, respectively, indicating
that groups of harder examples have longer mean reasoning spans, while
reasoning remains short in every group. The teacher increases the
update magnitudes for tokens in student trajectories that describe
anchors and spatial relations, helping short reasoning chains retain
the necessary information.

\begin{table}[t]
  \centering
  \caption{ReasonSeg test-set results stratified by query type and difficulty. Each entry is gIoU (\%) $\uparrow$ / \#Tok $\downarrow$. Bold indicates the best value for each metric in each stratum. \#Tok denotes the mean number of tokens within successfully parsed, closed \texttt{<think>} spans.}
  \label{tab:reasonseg_stratified}
  \fontsize{8}{9.2}\selectfont
  \renewcommand{\arraystretch}{1.08}
  \setlength{\tabcolsep}{2pt}
  \begin{tabular*}{\columnwidth}{@{\extracolsep{\fill}}lrccc@{}}
    \toprule
    Stratum & $N$ & SegZero & DPAD & \textbf{HISD} \\
    \midrule
    \rowcolor{tableband}\multicolumn{5}{c}{\textit{Query type}} \\
    Attribute & 99  & 59.26 / 129.02 & 61.49 / 75.07 & \textbf{62.79} / \textbf{11.51} \\
    Relation  & 48  & \textbf{59.90} / 132.83 & 55.96 / 76.46 & 59.26 / \textbf{11.04} \\
    Logic     & 632 & 55.63 / 123.80 & 56.93 / 79.42 & \textbf{60.18} / \textbf{14.70} \\
    \midrule
    \rowcolor{tableband}\multicolumn{5}{c}{\textit{Difficulty}} \\
    Easy      & 173 & 67.49 / 122.46 & 68.32 / 75.01 & \textbf{73.10} / \textbf{13.24} \\
    Medium    & 417 & 57.48 / 125.72 & 58.12 / 78.57 & \textbf{60.64} / \textbf{13.76} \\
    Hard      & 189 & 43.67 / 125.92 & 46.03 / 82.33 & \textbf{48.47} / \textbf{15.52} \\
    \bottomrule
  \end{tabular*}
\end{table}

\paragraph{CLIP alignment diagnostic.}
We use a frozen CLIP ViT-B/32 to compare anchored descriptions and reasoning
text with ground-truth bounding-box crops (ROI) and full images (AOI).
Table~\ref{tab:clip-discriminative-alignment} shows that HISD uses
80.17--91.83\% fewer reasoning tokens than DPAD across the five splits. Its
ROI/AOI ratios for both anchored descriptions and reasoning text are higher
on every split, while its absolute ROI similarities are lower. This is
consistent with HISD's design: after observing the target, cues, and
distractors, the teacher favors tokens describing the target and its key
attributes in student trajectories, increasing their relative update
magnitudes on positive-advantage trajectories. The resulting reasoning
chains are shorter and more focused on the target.

\begin{table}[t]
  \centering
  \caption{CLIP alignment and reasoning length. Shown are ROI similarities $S_1$ and $TS_1$ and the means of per-sample ROI/AOI ratios, SNR and TSNR; full-image means $S_2$ and $TS_2$ are omitted. NA: missing anchored descriptions. Bold: shortest reasoning and highest TSNR per split.}
  \label{tab:clip-discriminative-alignment}
  \fontsize{8}{9.2}\selectfont
  \renewcommand{\arraystretch}{1.08}
  \setlength{\tabcolsep}{1.8pt}
  \begin{tabular*}{\columnwidth}{@{\extracolsep{\fill}}clrrrrr@{}}
    \toprule
    Split & Method & \#Tok $\downarrow$ & $S_1$ & SNR & $TS_1$ & TSNR \\
    \midrule
    \multirow[c]{3}{*}{ReasonSeg test} & SegZero & 125.04 & NA & NA & 28.92 & 0.93 \\
                   & DPAD & 78.69 & 28.27 & 1.13 & 29.09 & 0.99 \\
                   & \textbf{HISD} & \textbf{14.07} & 28.15 & 1.16 & 27.25 & \textbf{1.02} \\
    \midrule
    \multirow[c]{3}{*}{ReasonSeg val} & SegZero & 126.95 & NA & NA & 29.84 & 0.93 \\
                   & DPAD & 77.36 & 29.56 & 1.14 & 29.83 & 1.00 \\
                   & \textbf{HISD} & \textbf{15.34} & 29.01 & 1.16 & 27.71 & \textbf{1.04} \\
    \midrule
    \multirow[c]{3}{*}{RefCOCO testA} & SegZero & 135.14 & NA & NA & 29.46 & 0.91 \\
                   & DPAD & 69.26 & 30.44 & 1.32 & 29.16 & 1.08 \\
                   & \textbf{HISD} & \textbf{5.66} & 30.07 & 1.35 & 25.36 & \textbf{1.10} \\
    \midrule
    \multirow[c]{3}{*}{RefCOCO+ testA} & SegZero & 135.28 & NA & NA & 29.49 & 0.91 \\
                   & DPAD & 70.55 & 30.15 & 1.32 & 29.35 & 1.08 \\
                   & \textbf{HISD} & \textbf{6.00} & 29.93 & 1.36 & 26.18 & \textbf{1.16} \\
    \midrule
    \multirow[c]{3}{*}{RefCOCOg test} & SegZero & 130.52 & NA & NA & 30.46 & 0.94 \\
                   & DPAD & 71.75 & 30.77 & 1.24 & 30.61 & 1.06 \\
                   & \textbf{HISD} & \textbf{9.18} & 30.63 & 1.27 & 28.78 & \textbf{1.11} \\
    \bottomrule
  \end{tabular*}
\end{table}

\begingroup
\interlinepenalty=10000
\paragraph{Qualitative analysis.}
\label{sec:qualitative_analysis}
Figure~\ref{fig:qualitative} compares SegZero, DPAD, and HISD on three
reasoning segmentation queries. The excerpts show how the target is
interpreted, while the masks and per-sample IoU show how accurately it is
localized. Token counts refer to the lengths of full reasoning spans.
\par
\endgroup

For ``the hardest body part of the animal''
(Fig.~\ref{fig:qualitative}, horns example), HISD identifies the horns and
achieves 95.24\% IoU with 22 tokens. SegZero also names the horns but obtains
54.91\% IoU with 104 tokens; DPAD obtains 55.93\% with 73 tokens. Mentioning
the correct noun therefore does not guarantee accurate localization. For
the insect-camouflage query (Fig.~\ref{fig:qualitative}, leaves example),
both baselines localize the insect rather than the queried plant part,
yielding 0\% IoU with 159 and 96 tokens, respectively. HISD explicitly
identifies the leaves and reaches 95.92\% IoU with 12 tokens. For the
aircraft-takeoff query (Fig.~\ref{fig:qualitative}, runway example), HISD
selects the runway, achieving 95.44\% IoU with 16 tokens. SegZero and DPAD
instead localize the aircraft, using 168 and 97 tokens, respectively, and
each scoring 0.03\% IoU. These examples show that HISD's outputs provide
concise target disambiguation and illustrate the correspondence between
this disambiguation and accurate localization.

\begin{figure*}[t]
  \begin{minipage}[t]{\columnwidth}
  \vspace{0pt}
  \centering
  \includegraphics[width=\linewidth]{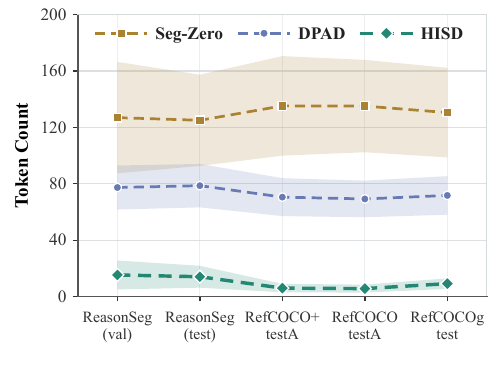}
  \caption{Reasoning-token counts across five evaluation splits. Lines show means; shaded bands indicate the mean $\pm$ one population standard deviation computed from each method's valid, closed, parsed reasoning spans.}
  \label{fig:token_count}
  \end{minipage}\hfill
  \begin{minipage}[t]{\columnwidth}
  \vspace{0pt}
  \input{sec/figures/5_token_scatter}
  \end{minipage}
\end{figure*}

Figure~\ref{fig:token_count} extends the reasoning-length comparison to five
evaluation splits. On every split, HISD has a lower mean token count
(5.66--15.34 tokens) and smaller absolute standard deviation
(2.82--10.26 tokens) than SegZero and DPAD. The shaded bands describe
variability across each method's valid samples. These results show that
HISD strikes a favorable balance between visual reasoning for localization
and accurate segmentation and has advantages in both reasoning and
referring expression segmentation.

Figure~\ref{fig:token_scatter} compares reasoning-token counts for complete
three-model paired samples from ReasonSeg test. HISD uses fewer reasoning
tokens on every paired sample, averaging 14.10 tokens versus 78.73 for DPAD
and 125.04 for SegZero.

\subsection{Ablation Studies}
\label{sec:ablation_studies}

We ablate three HISD design choices: teacher updates, privileged inputs, and
token reweighting. Each group changes only one factor while keeping all
other settings fixed.

\begin{table*}[p]
  \centering
  \small
  \renewcommand{\arraystretch}{1.23}
  \setlength{\tabcolsep}{3pt}
  \caption{HISD ablations with Qwen2.5-VL-7B-Instruct. Panels vary (a) teacher updates, (b) privileged inputs, and (c) token reweighting. Entries are gIoU / cIoU (\%); bold indicates the best result in each panel.}
  \label{tab:hisd_ablation}
  \label{tab:teacher-update-ablation}
  \label{tab:ema_teacher_input_ablation}
  \label{tab:ema_teacher_token_ablation}
  \begin{tabular*}{\linewidth}{@{\extracolsep{\fill}}l*{6}{c}@{}}
    \toprule
    Variant & \shortstack{ReasonSeg\\test}
      & \shortstack{ReasonSeg\\val}
      & \shortstack{RefCOCO\\testA}
      & \shortstack{RefCOCO+\\testA}
      & \shortstack{RefCOCOg\\test} & Avg. \\
    \midrule
    \rowcolor{tableband}\multicolumn{7}{c}{\textit{(a) Teacher updates}} \\
    Current policy (every step)
      & 58.52 / 53.64 & 61.80 / 53.87 & 80.33 / 78.98
      & 77.44 / 75.39 & 74.79 / 72.66 & 70.58 / 66.91 \\
    Hard update (every 10 steps)
      & 54.64 / 48.23 & 58.92 / 49.23 & 80.40 / 79.46
      & 77.39 / 75.04 & 74.30 / 72.18 & 69.13 / 64.83 \\
    Frozen policy
      & 54.51 / 49.12 & 60.21 / 54.79 & 80.79 / 79.74
      & 77.29 / 75.14 & 74.40 / 72.40 & 69.44 / 66.24 \\
    \textbf{EMA (Ours)}
      & \textbf{60.45 / 57.49} & \textbf{63.76 / 61.98}
      & \textbf{80.96 / 80.10} & \textbf{78.19 / 76.31}
      & \textbf{75.02 / 73.24} & \textbf{71.68 / 69.83} \\
    \midrule
    \rowcolor{tableband}\multicolumn{7}{c}{\textit{(b) Privileged inputs}} \\
    No privileged information
      & 57.45 / 53.94 & 60.45 / 57.38 & 79.85 / 79.12
      & 76.78 / 75.02 & 73.45 / 71.24 & 69.60 / 67.34 \\
    Answer-only
      & 59.69 / 56.57 & 60.99 / 59.39 & 80.10 / 78.89
      & 76.80 / 74.50 & 74.13 / 71.95 & 70.34 / 68.26 \\
    Gaussian blur
      & 56.17 / 52.13 & 62.08 / 57.32 & 80.59 / 79.35
      & 77.10 / 75.09 & 74.12 / 72.12 & 70.01 / 67.20 \\
    \textbf{Segment Hint (Ours)}
      & \textbf{60.45 / 57.49} & \textbf{63.76 / 61.98}
      & \textbf{80.96 / 80.10} & \textbf{78.19 / 76.31}
      & \textbf{75.02 / 73.24} & \textbf{71.68 / 69.83} \\
    \midrule
    \rowcolor{tableband}\multicolumn{7}{c}{\textit{(c) Token reweighting}} \\
    Top-$K$, positive + negative
      & 57.88 / 53.13 & 62.35 / 57.17 & 80.58 / 79.01
      & 77.04 / 75.03 & 74.01 / 71.47 & 70.37 / 67.16 \\
    Sampled token, positive only
      & 57.73 / 54.09 & 62.30 / 56.67 & 80.52 / 79.23
      & 77.00 / 74.81 & 74.48 / 72.28 & 70.41 / 67.41 \\
    \textbf{Top-$K$, positive only (Ours)}
      & \textbf{60.45 / 57.49} & \textbf{63.76 / 61.98}
      & \textbf{80.96 / 80.10} & \textbf{78.19 / 76.31}
      & \textbf{75.02 / 73.24} & \textbf{71.68 / 69.83} \\
    \bottomrule
  \end{tabular*}
\end{table*}

\begingroup
\widowpenalty=10000
\paragraph{Teacher updates.}
With Segment Hints and positive-only Top-$K$ reweighting fixed, we compare
using the current student parameters at every step, hard synchronization
every 10 optimizer steps with freezing between updates, a teacher frozen
after initialization, and stepwise EMA (Eq.~\eqref{eq:ema_update}).
All teachers rescore complete student trajectories without computing
gradients. Relative to EMA, the first three variants reduce cIoU on
ReasonSeg validation by
8.11, 12.75, and 7.19 percentage points, respectively
(Table~\ref{tab:teacher-update-ablation}(a)). By comparison, their cIoU gaps
on the three RefCOCO-family splits are at most 1.27 percentage points.
These results support the use of EMA among the update schemes tested under
the current configuration, especially on ReasonSeg, where EMA's parameter
averaging can provide smoother teacher scores.
\endgroup

\paragraph{Privileged inputs.}
With EMA and positive-only Top-$K$ reweighting fixed, we compare no
privileged information, the ground-truth bounding-box
answer alone, a visual input with Gaussian blur applied outside the
ground-truth box, and Segment Hints. Compared with no privileged information,
Segment Hints improve gIoU/cIoU by 3.00/3.55 percentage points on ReasonSeg
test and by 3.31/4.60 percentage points on validation
(Table~\ref{tab:ema_teacher_input_ablation}(b)). Answer-only inputs improve
performance on ReasonSeg but reduce cIoU on RefCOCO testA and RefCOCO+ testA
by 0.23 and 0.52 percentage points, respectively; Gaussian blur reduces both
metrics on ReasonSeg test. A teacher given the ground-truth box can favor
coordinate tokens consistent with the true location but struggles to
determine whether reasoning tokens capture the correct discriminative
cues. Without these cues, the teacher signal acts more like a localization
shortcut. Gaussian blur also lowers the average cIoU below that of the
no-privileged-information variant. Because this variant changes the
teacher's visual input, $\Delta_t$ also incorporates perceptual differences
caused by image perturbations unrelated to target disambiguation.

\begingroup
\brokenpenalty=10000
\paragraph{Token reweighting.}
As shown in Table~\ref{tab:ema_teacher_token_ablation}(c), both modifications
score below full HISD on both metrics across all five splits. Compared with
sampled-token reweighting, full HISD improves cIoU by 3.40 and 5.31
percentage points on ReasonSeg test and validation, respectively; compared
with reweighting both positive and negative advantages, it improves cIoU by
4.36 and 4.81 percentage points, respectively. These two modifications
reduce average cIoU by 2.42 and 2.67 percentage points, respectively. Both
variants also use Segment Hints, showing that Segment Hints do not
automatically improve performance: privileged information is effectively
used only when paired with an appropriate reweighting scheme.
\endgroup

\section{Conclusion}
\label{sec:conclusion}

HISD links query-relevant visual evidence to selective token-level credit
modulation for reasoning segmentation. Based on the evidence-ratio principle,
it uses annotation-derived Segment Hints to modulate positive group-relative
advantages. Compared with our reproduction of DPAD~\cite{ref3}, the evaluated
model improves segmentation accuracy while generating shorter valid reasoning
spans. In the reported ablations, Segment Hints outperform
bounding-box-only answers and inputs with Gaussian blur outside the
bounding box, and Top-$K$ reweighting of positive advantages only also
outperforms the two tested weighting variants. These comparisons support
matching teacher evidence to its use in optimization under the current
evaluation setting. The independent contributions of individual hint fields,
as well as generalization to other generators, model sizes, and data domains,
remain to be evaluated.

\clearpage
\raggedbottom
\ArxivResetColumnHeight
\pdfbookmark[0]{LLM Usage Statement}{llm-usage-statement}
\section*{LLM Usage Statement}
ChatGPT 6.1 SOL was used for English-language polishing throughout the
manuscript. GPT-based assistants also helped with figure preparation and
LaTeX formatting. In the research pipeline, Gemini 2.5 Pro generated
structured Segment Hints offline from training images, queries, and target
annotations, and a frozen Qwen2.5-VL-72B classifier assigned query types
for stratified evaluation, as described in the paper.
The authors retain full responsibility for all scientific claims,
experimental results, references, and the final manuscript.

{
  \small
  \setlength{\bibsep}{0pt}
  \bibliographystyle{ieeenat_fullname}
  \bibliography{main}

\begin{thebibliography}{25}
\providecommand{\natexlab}[1]{#1}
\providecommand{\url}[1]{\texttt{#1}}
\expandafter\ifx\csname urlstyle\endcsname\relax
  \providecommand{\doi}[1]{doi: #1}\else
  \providecommand{\doi}{doi: \begingroup \urlstyle{rm}\Url}\fi

\bibitem[Bai et~al.(2025)Bai, Chen, Liu, et~al.]{ref43}
S. Bai, K. Chen, X. Liu, et~al.
\newblock {Qwen2.5-VL} technical report, 2025.
\newblock arXiv:2502.13923.

\bibitem[Chen et~al.(2024)Chen, Li, Sun, Wang, and Chen]{sam4mllm}
Yi-Chia Chen, Wei-Hua Li, Cheng Sun, Yu-Chiang~Frank Wang, and Chu-Song Chen.
\newblock {SAM4MLLM}: Enhance multi-modal large language model for referring
  expression segmentation.
\newblock In \emph{Computer Vision -- ECCV 2024}, pages 323--340. Springer,
  2024.

\bibitem[Chng et~al.(2024)Chng, Zheng, Han, Qiu, and Huang]{magnet}
Yong~Xien Chng, Henry Zheng, Yizeng Han, Xuchong Qiu, and Gao Huang.
\newblock Mask grounding for referring image segmentation.
\newblock In \emph{Proceedings of the IEEE/CVF Conference on Computer Vision
  and Pattern Recognition (CVPR)}, pages 26573--26583, 2024.

\bibitem[Comanici et~al.(2025)Comanici, Bieber, Schaekermann, et~al.]{ref46}
G. Comanici, E. Bieber, M. Schaekermann, et~al.
\newblock {Gemini 2.5}: Pushing the frontier with advanced reasoning,
  multimodality, long context, and next generation agentic capabilities, 2025.
\newblock arXiv:2507.06261.

\bibitem[Huang et~al.(2025)Huang, Xu, Zhou, et~al.]{ref5}
J. Huang, Z. Xu, J. Zhou, et~al.
\newblock {SAM-R1}: Leveraging {SAM} for reward feedback in multimodal
  segmentation via reinforcement learning.
\newblock In \emph{Advances in Neural Information Processing Systems}, 2025.

\bibitem[H{\"u}botter et~al.(2026)H{\"u}botter, L{\"u}beck, Behric,
  et~al.]{ref31}
J. H{\"u}botter, F. L{\"u}beck, L. Behric, et~al.
\newblock Reinforcement learning via self-distillation, 2026.
\newblock arXiv:2601.20802.

\bibitem[Lai et~al.(2024)Lai, Tian, Chen, et~al.]{ref1}
X. Lai, Z. Tian, Y. Chen, et~al.
\newblock {LISA}: Reasoning segmentation via large language model.
\newblock In \emph{Proceedings of the IEEE/CVF Conference on Computer Vision
  and Pattern Recognition}, 2024.

\bibitem[Liang et~al.(2023)Liang, Wu, Dai, Li, Zhao, Zhang, Zhang, Vajda, and
  Marculescu]{ovseg}
Feng Liang, Bichen Wu, Xiaoliang Dai, Kunpeng Li, Yinan Zhao, Hang Zhang,
  Peizhao Zhang, Peter Vajda, and Diana Marculescu.
\newblock Open-vocabulary semantic segmentation with mask-adapted {CLIP}.
\newblock In \emph{Proceedings of the IEEE/CVF Conference on Computer Vision
  and Pattern Recognition (CVPR)}, pages 7061--7070, 2023.

\bibitem[Liu et~al.(2023)Liu, Ding, and Jiang]{gres}
Chang Liu, Henghui Ding, and Xudong Jiang.
\newblock {GRES}: Generalized referring expression segmentation.
\newblock In \emph{Proceedings of the IEEE/CVF Conference on Computer Vision
  and Pattern Recognition (CVPR)}, pages 23592--23601, 2023.

\bibitem[Liu et~al.(2025{\natexlab{a}})Liu, Peng, Zhong, et~al.]{ref2}
Y. Liu, B. Peng, Z. Zhong, et~al.
\newblock {Seg-Zero}: Reasoning-chain guided segmentation via cognitive
  reinforcement, 2025{\natexlab{a}}.
\newblock arXiv:2503.06520.

\bibitem[Liu et~al.(2025{\natexlab{b}})Liu, Qu, Zhong, et~al.]{ref4}
Y. Liu, T. Qu, Z. Zhong, et~al.
\newblock {VisionReasoner}: Unified reasoning-integrated visual perception via
  reinforcement learning, 2025{\natexlab{b}}.
\newblock arXiv:2505.12081.

\bibitem[Mao et~al.(2016)Mao, Huang, Toshev, et~al.]{ref45}
J. Mao, J. Huang, A. Toshev, et~al.
\newblock Generation and comprehension of unambiguous object descriptions.
\newblock In \emph{Proceedings of the IEEE Conference on Computer Vision and
  Pattern Recognition}, 2016.

\bibitem[Radford et~al.(2021)Radford, Kim, Hallacy, et~al.]{ref44}
A. Radford, J.~W. Kim, C. Hallacy, et~al.
\newblock Learning transferable visual models from natural language
  supervision, 2021.
\newblock arXiv:2103.00020.

\bibitem[Ravi et~al.(2024)Ravi, Gabeur, Hu, et~al.]{ref42}
N. Ravi, V. Gabeur, Y.~T. Hu, et~al.
\newblock {SAM 2}: Segment anything in images and videos, 2024.
\newblock arXiv:2408.00714.

\bibitem[Ren et~al.(2024{\natexlab{a}})Ren, Liu, Zeng, Lin, Li, Cao, Chen,
  Huang, Chen, Yan, Zeng, Zhang, Li, Yang, Li, Jiang, and Zhang]{grounded_sam}
Tianhe Ren, Shilong Liu, Ailing Zeng, Jing Lin, Kunchang Li, He Cao, Jiayu
  Chen, Xinyu Huang, Yukang Chen, Feng Yan, Zhaoyang Zeng, Hao Zhang, Feng Li,
  Jie Yang, Hongyang Li, Qing Jiang, and Lei Zhang.
\newblock {Grounded SAM}: Assembling open-world models for diverse visual
  tasks, 2024{\natexlab{a}}.

\bibitem[Ren et~al.(2024{\natexlab{b}})Ren, Huang, Wei, Zhao, Fu, Feng, and
  Jin]{pixellm}
Zhongwei Ren, Zhicheng Huang, Yunchao Wei, Yao Zhao, Dongmei Fu, Jiashi Feng,
  and Xiaojie Jin.
\newblock {PixelLM}: Pixel reasoning with large multimodal model.
\newblock In \emph{Proceedings of the IEEE/CVF Conference on Computer Vision
  and Pattern Recognition (CVPR)}, pages 26374--26383, 2024{\natexlab{b}}.

\bibitem[Shao et~al.(2024)Shao, Wang, Zhu, et~al.]{ref14}
Z. Shao, P. Wang, Q. Zhu, et~al.
\newblock {DeepSeekMath}: Pushing the limits of mathematical reasoning in open
  language models, 2024.
\newblock arXiv:2402.03300.

\bibitem[Tarvainen and Valpola(2017)]{ref28}
A. Tarvainen and H. Valpola.
\newblock Mean teachers are better role models: Weight-averaged consistency
  targets improve semi-supervised deep learning results.
\newblock In \emph{Advances in Neural Information Processing Systems}, 2017.

\bibitem[Wang et~al.(2025)Wang, Fang, Kong, Li, Xu, Yang, Li, Zhu, and
  Wang]{pixelthink}
Song Wang, Gongfan Fang, Lingdong Kong, Xiangtai Li, Jianyun Xu, Sheng Yang,
  Qiang Li, Jianke Zhu, and Xinchao Wang.
\newblock {PixelThink}: Towards efficient chain-of-pixel reasoning, 2025.

\bibitem[Yan et~al.(2026)Yan, Qiu, Zhang, et~al.]{ref33}
C. Yan, Y. Qiu, W. Zhang, et~al.
\newblock Revisit to segment: Working memory distillation for reasoning
  segmentation, 2026.
\newblock arXiv:2609.34863.

\bibitem[Yang et~al.(2026{\natexlab{a}})Yang, Qin, Si, et~al.]{ref32}
C. Yang, C. Qin, Q. Si, et~al.
\newblock Self-distilled {RLVR}, 2026{\natexlab{a}}.
\newblock arXiv:2604.03128.

\bibitem[Yang et~al.(2026{\natexlab{b}})Yang, Zhou, Li, et~al.]{ref3}
T. Yang, Q. Zhou, Y. Li, et~al.
\newblock Discriminative perception via anchored description for reasoning
  segmentation.
\newblock In \emph{Proceedings of the IEEE/CVF Conference on Computer Vision
  and Pattern Recognition}, 2026{\natexlab{b}}.

\bibitem[Yang et~al.(2022)Yang, Wang, Tang, Chen, Zhao, and Torr]{lavt}
Zhao Yang, Jiaqi Wang, Yansong Tang, Kai Chen, Hengshuang Zhao, and Philip~H.S.
  Torr.
\newblock {LAVT}: Language-aware vision transformer for referring image
  segmentation.
\newblock In \emph{Proceedings of the IEEE/CVF Conference on Computer Vision
  and Pattern Recognition (CVPR)}, pages 18155--18165, 2022.

\bibitem[You and Wu(2025)]{ref6}
Z. You and Z. Wu.
\newblock {Seg-R1}: Segmentation can be surprisingly simple with reinforcement
  learning, 2025.
\newblock arXiv:2506.22624.

\bibitem[Zhao et~al.(2026)Zhao, Xie, Liu, et~al.]{ref29}
S. Zhao, Z. Xie, M. Liu, et~al.
\newblock Self-distilled reasoner: On-policy self-distillation for large
  language models, 2026.
\newblock arXiv:2601.18734.

\end{thebibliography}
}

\ifArxivIncludeSupplement
  \raggedbottom
\clearpage
\onecolumn
\nolinenumbers
\setlength{\multicolsep}{0pt}
\setcounter{page}{1}
\renewcommand{\thepage}{S\arabic{page}}
\begingroup
\centering
\Large
\textbf{\thetitle}\par
\vspace{0.4em}Supplementary Material\par
\vspace{0.5em}
\endgroup
\pdfbookmark[0]{Supplementary Material}{supplementary-material}
\phantomsection
\label{sec:supplementary_start}
\appendix
\renewcommand{\thetable}{\thesection\arabic{table}}
\renewcommand{\theHtable}{\thesection.\arabic{table}}
\makeatletter
\setlength{\@fptop}{0pt}
\setlength{\@fpsep}{12pt}
\setlength{\@fpbot}{0pt plus 1fil}
\setlength{\@dblfptop}{0pt}
\setlength{\@dblfpsep}{12pt}
\setlength{\@dblfpbot}{0pt plus 1fil}
\makeatother
\begin{multicols}{2}
\fontsize{9}{9.5}\selectfont
\section{Implementation Details}
\label{sec:implementation_appendix}
\setcounter{table}{0}

\subsection{Training Procedure and Hyperparameters}
\label{sec:training_details}
\noindent\textbf{Algorithm 1: One-step HISD training procedure}
\begingroup
\fontsize{9}{9.5}\selectfont
\begin{enumerate}[leftmargin=*,itemsep=1.5pt,topsep=2.5pt]

\setcounter{enumi}{0}
\item Before training begins, use Gemini 2.5 Pro~\cite{ref46} offline to
generate a structured Segment Hint $H$ once for each training sample
$(I,Q,B^{\mathrm{gt}})$ according to Eq.~\eqref{eq:hint_generation}; use
it as the fixed teacher context for that sample.
\item For each $x=(I,Q)$ in the batch, sample $G=8$ responses
$y^{(i)}$ from $\pi_{\theta_{\mathrm{old}}}$.
\item For each response, the frozen SAM2-Large~\cite{ref42}
deterministically generates a mask from its predicted localization set;
compute $R(x,y^{(i)})$.
\item Normalize rewards within each group according to
Eq.~\eqref{eq:trajectory_advantage} to obtain trajectory-level
advantages $A^{(i)}$.
\item The exponential moving average (EMA) teacher replays and scores each
completed trajectory under $(I,Q\oplus H)$ without gradients, while the
student performs one no-gradient signal forward pass under $(I,Q)$.
Use the union of the teacher's Top-$K$ tokens and the sampled token
$y_t^{(i)}$ as the restricted support set $C_t$. Compute the
teacher--student endorsement gap $\Delta_t^{(i)}$ and reweighting factor
$m_t^{(i)}$ according to Eqs.~\eqref{eq:teacher_student_gap}
and~\eqref{eq:token_weight}.
\item Compute the current $\lambda_s$ according to
Eq.~\eqref{eq:curriculum_schedule}, and obtain the corrected advantage
$\hat{A}_t^{(i)}$ according to Eq.~\eqref{eq:corrected_advantage}: apply
teacher reweighting to trajectories with $A^{(i)}>0$, and retain the
original advantage for trajectories with $A^{(i)}\leq 0$.
\item Perform a fresh gradient-enabled student forward pass, compute the
loss $\mathcal{L}_{\mathrm{HISD}}(\theta)
=-\frac{1}{G}\sum_{i=1}^{G}\frac{1}{|y^{(i)}|}
\sum_{t=1}^{|y^{(i)}|}\ell_t^{(i)}(\theta)$ according to
Eq.~\eqref{eq:grpo_token_objective}, and backpropagate only to the
student parameters $\theta$.
\item Take an optimizer step (learning rate $1\times10^{-6}$ and weight
decay $0.01$). After the step, apply the EMA update in
Eq.~\eqref{eq:ema_update}:
$\bar{\theta}\leftarrow0.95\bar{\theta}+0.05\theta$.
\item The value of $\lambda_s$ automatically decays with the number of
optimizer steps according to Eq.~\eqref{eq:curriculum_schedule}.
Proceed to the next optimizer step and repeat Steps 2--8.
\end{enumerate}
\endgroup

\section{Additional Efficiency Analysis}
\label{sec:efficiency_appendix}
\setcounter{table}{0}

\paragraph{Equal-budget replay.}
For each data split, we set a shared response-token budget $K$ by rounding
HISD's mean native response length after retokenization.
We truncate the existing responses to $K$ tokens and reapply each method's
parser, assigning zero IoU to samples with parsing failures.
Thus, Table~\ref{tab:equal-token-budget} estimates performance under truncation;
it does not evaluate the results of re-decoding with budget constraints
imposed during generation.
The budget covers the full response, whereas the reasoning-length analysis
in Section~\ref{sec:main_results} counts only closed reasoning text.
At $K=98$--106, HISD retains parseable predictions on 99.50--100.00\% of samples,
with a gIoU decrease of at most 0.11 points.
DPAD's retention rate is 1.62--2.50\%, whereas SegZero retains no samples
under its parser.

\paragraph{Pareto comparison.}
We compute the frontier over the scanned replay points using the mean token
count of retained responses and gIoU: a point is dominated if another point
uses no more tokens and attains no lower gIoU, with at least one strict
improvement.
Across all five data splits, every nonzero frontier point belongs to HISD.
On ReasonSeg test, HISD reaches 50.40 gIoU at $K=80$, 58.38 at $K=88$,
and its native level of 60.45 at $K=112$; on RefCOCO testA, it reaches its
native level of 80.96 at $K=88$.
\end{multicols}

\noindent\begin{minipage}[b]{0.48\textwidth}
  \centering
  \renewcommand{\thetable}{A\arabic{table}}
  \renewcommand{\theHtable}{A.\arabic{table}}
  \setcounter{table}{0}
  \captionsetup{font=small,skip=4pt,hypcap=false}
  \captionof{table}{Summary of HISD training hyperparameters.}
  \label{tab:training_hyperparameters}
  \fontsize{9}{10.5}\selectfont
  \setlength{\tabcolsep}{2.2pt}
  \renewcommand{\arraystretch}{1.25}
  \begin{tabular}{@{}p{0.34\columnwidth}p{\dimexpr0.66\columnwidth-2\tabcolsep\relax}@{}}
    \toprule
    Item & Value \\
    \midrule
    Vision--language policy backbone
      & Qwen2.5-VL~\cite{ref43} (trainable) \\
    Segmentation module
      & SAM2-Large~\cite{ref42}, frozen during training \\
    Discriminative reward metric
      & CLIP ViT-B/32~\cite{ref44}, frozen \\
    Training data
      & 3,000 RefCOCOg~\cite{ref45} samples \\
    Group size / batch size
      & $G=8$; batch size $16$ \\
    Optimizer
      & learning rate $1\times10^{-6}$; weight decay $0.01$ \\
    Restricted support size
      & $K=16$ \\
    Symmetric token-reweighting clip
      & $\epsilon_w=0.1$ (range $[0.9,1.1]$) \\
    Curriculum-decay initial value / window
      & $\lambda_0=0.5$; $\kappa=0.25$ (linear decay to $0$ over the first $25\%$ of all optimizer steps)
      \\
    EMA student mixing rate
      & $\beta=0.05$ (teacher retention $0.95$) \\
    \bottomrule
  \end{tabular}
\end{minipage}\hfill
\begin{minipage}[b]{0.48\textwidth}
  \centering
  \renewcommand{\thetable}{B\arabic{table}}
  \renewcommand{\theHtable}{B.\arabic{table}}
  \setcounter{table}{0}
  \captionsetup{font=footnotesize,skip=4pt,hypcap=false}
  \captionof{table}{Equal-response-token-budget replay.
  Each cell shows the native result $\to$ the replay result at $K$;
  native decoding uses \texttt{max\_new\_tokens=1024}.
  Retained is the percentage of samples with a parseable prediction.
  Mean token counts measure the full response actually retained.
  gIoU and retention fractions are expressed in \%; samples with parsing
  failures are assigned zero IoU.
  Bold indicates the best gIoU before and after truncation.}
  \label{tab:equal-token-budget}
  \fontsize{8}{9.2}\selectfont
  \renewcommand{\arraystretch}{1.05}
  \setlength{\tabcolsep}{2.2pt}
  \begin{tabular*}{\columnwidth}{@{\extracolsep{\fill}}lrrr@{}}
    \toprule
    Metric & SegZero & DPAD & \textbf{HISD} \\
    \midrule
    \multicolumn{4}{@{}l}{\textit{ReasonSeg test} ($K=105$)} \\
    Mean tokens & $211.14\to105.00$ & $166.55\to105.00$ & $105.28\to100.52$ \\
    gIoU & $56.36\to0.00$ & $57.45\to1.27$ & $\mathbf{60.45}\to\mathbf{60.34}$ \\
    Retained & $97.82\to0.00$ & $100.00\to1.93$ & $100.00\to99.61$ \\
    \midrule
    \multicolumn{4}{@{}l}{\textit{ReasonSeg val} ($K=106$)} \\
    gIoU & $60.05\to0.00$ & $60.45\to1.53$ & $\mathbf{63.76}\to\mathbf{63.76}$ \\
    Retained & $99.00\to0.00$ & $100.00\to2.50$ & $100.00\to99.50$ \\
    \midrule
    \multicolumn{4}{@{}l}{\textit{RefCOCO testA} ($K=99$)} \\
    gIoU & $74.39\to0.00$ & $79.85\to1.57$ & $\mathbf{80.96}\to\mathbf{80.96}$ \\
    Retained & $93.01\to0.00$ & $99.95\to1.92$ & $99.95\to99.95$ \\
    \midrule
    \multicolumn{4}{@{}l}{\textit{RefCOCO+ testA} ($K=98$)} \\
    gIoU & $72.34\to0.00$ & $76.78\to1.31$ & $\mathbf{78.19}\to\mathbf{78.19}$ \\
    Retained & $93.72\to0.00$ & $99.90\to1.62$ & $100.00\to100.00$ \\
    \midrule
    \multicolumn{4}{@{}l}{\textit{RefCOCOg test} ($K=102$)} \\
    gIoU & $70.56\to0.00$ & $73.45\to1.98$ & $\mathbf{75.02}\to\mathbf{75.02}$ \\
    Retained & $95.56\to0.00$ & $99.98\to2.45$ & $100.00\to100.00$ \\
    \bottomrule
  \end{tabular*}
\end{minipage}

\onecolumn
\section{Offline Segment Hint Generation Prompt}
\label{sec:hint_generation_prompt}
\begin{segmenthintprompt}
You are a visual-grounding analyst with access to privileged information. This request contains two complete views of the same scene. Image 1 is the unmodified original scene. Image 2 is an audit copy in which the intended segmentation target is enclosed by a thin red rectangle, and two positive interior locations on that same target are marked by a lime point and a cyan point, respectively.

\textbf{Original referring expression:}\\
\texttt{\{\{REFERRING\_EXPRESSION\_JSON\}\}}

\par\noindent{\color{black!50}\rule{\linewidth}{0.4pt}}\par
Use the red rectangle and both positive points only to identify the intended object. Determine \texttt{target} from the visible object in the images together with the referring expression; no pre-extracted or fixed category label is provided. Do not merely copy a noun from the referring expression when the visible category is different. Return a short, commonly used object-category name supported by the image. Describe naturally visible evidence from Image 1, never the artificial annotations themselves. The result will be included in the teacher prompt, so every item must be concise, image-grounded, and useful for distinguishing the target from visually confusable objects.

Attribute ownership is critical in crowded or overlapping scenes. Both positive points belong to the intended target, but the red rectangle may also contain background or neighboring objects. Do not copy clothing, accessories, colors, parts, poses, or other attributes from neighboring objects. If an attribute is occluded or uncertain, omit it.

Return exactly one JSON object containing only the following keys in this order, with no Markdown or extra text:
\par\smallskip\noindent\begin{minipage}{\linewidth}\ttfamily\raggedright
\{\par
"target": "\textless{}short object category inferred from the visible target\textgreater{}",\par
"cues": ["\textless{}visible, discriminative target cue\textgreater{}", "\textless{}visible, discriminative target cue\textgreater{}"],\par
"others": ["\textless{}confusable object or object group\textgreater{}: \textless{}key visible difference from the target\textgreater{}"],\par
"steps": ["\textless{}short, ordered localization instruction\textgreater{}"]\par
\}\par
\end{minipage}
\par\smallskip

\textbf{Requirements:}

\begin{itemize}[leftmargin=*,itemsep=4pt,topsep=4pt,parsep=0pt]
\item \texttt{target} must be inferred from the visible intended object; it is not supplied separately and must not be pre-extracted from the referring expression.
\item \texttt{cues} contains two to five visible target cues, emphasizing relative position, appearance, pose, scale, or occlusion.
\item \texttt{others} contains one to four strings. Every item must contain a colon separating the confusable object or object group from its key difference relative to the target.
\item \texttt{steps} contains one to five short, ordered localization instructions.
\item Do not mention either image number, the red rectangle, the colored points, annotations, bounding boxes, coordinates, pixels, or numerical geometry.
\item Do not output an answer box or point coordinates.
\item Use concise English strings and output JSON only.
\end{itemize}
\end{segmenthintprompt}

\fi
\end{document}